\documentclass[11pt,a4paper]{article}
\usepackage[hyperref]{emnlp2020}
\usepackage{times}
\usepackage{latexsym}

\usepackage{times}
\usepackage{latexsym}
\usepackage{url}
\usepackage{natbib}
\usepackage{graphicx}
\usepackage{comment}
\graphicspath{ {images/} }
\usepackage{amssymb}
\usepackage{amsmath}
\usepackage{bm}
\usepackage{enumitem}
\usepackage{url}
\usepackage{todonotes}
\usepackage{color}
\usepackage{tabularx}
\usepackage{url}
\usepackage[utf8]{inputenc}
\usepackage{caption}
\usepackage{subcaption}
\usepackage{color}
\usepackage[bottom]{footmisc}
\usepackage{textcomp}
\usepackage{comment}
\usepackage{multirow}
\usepackage{subcaption,siunitx,booktabs}
\usepackage{subcaption}
\usepackage{graphicx}
\usepackage{textcomp , xspace}
\usepackage{float}
\usepackage{appendix}
\usepackage{color,soul}
\usepackage{multirow}
\usepackage{subcaption}
\usepackage[skip=2.8 pt]{caption}

\usepackage{siunitx,tabularx,ragged2e,booktabs,caption}
\newcolumntype{Y}{>{\RaggedRight\arraybackslash}X}
\newcolumntype{T}[1]{S[table-format=#1]}

\makeatletter
\newcommand\xleftrightarrow[2][]{%
  \ext@arrow 9999{\longleftrightarrowfill@}{#1}{#2}}
\newcommand\longleftrightarrowfill@{%
  \arrowfill@\leftarrow\relbar\rightarrow}
\makeatother

\newcommand\Tstrut{\rule{0pt}{2.6ex}}         
\newcommand\Bstrut{\rule[-0.9ex]{0pt}{0pt}}   

\aclfinalcopy 
\def\aclpaperid{2808} 

\usepackage{microtype}

\aclfinalcopy 

\title{WNUT-2020 Task 1 Overview: Extracting Entities and Relations from Wet Lab Protocols}

\author{Jeniya Tabassum\textsuperscript{1}, Sydney Lee\textsuperscript{1}, Wei Xu\textsuperscript{2},  Alan Ritter\textsuperscript{2} \\ \textsuperscript{1} Department of Computer Science and Engineering, The Ohio State University\\
  \textsuperscript{2} School of Interactive Computing, Georgia Institute of Technology \\
  {\small \tt \{tabassum.13, lee.7509\}@osu.edu \quad \{wei.xu, alan.ritter\}@cc.gatech.edu }\\
}

\begin{document}

\maketitle

\begin{abstract}
This paper presents the results of the wet lab information extraction task at WNUT 2020. This task consisted of two sub tasks: (1) a Named Entity Recognition (NER) task with 13 participants and (2) a Relation Extraction (RE) task with 2 participants. We outline the task, data annotation process,  corpus statistics, and provide a high-level overview of the participating systems for each sub task.
\end{abstract}

\section{Introduction}

\begin{figure}[tb!]
    \centering
    \includegraphics[scale=0.17]{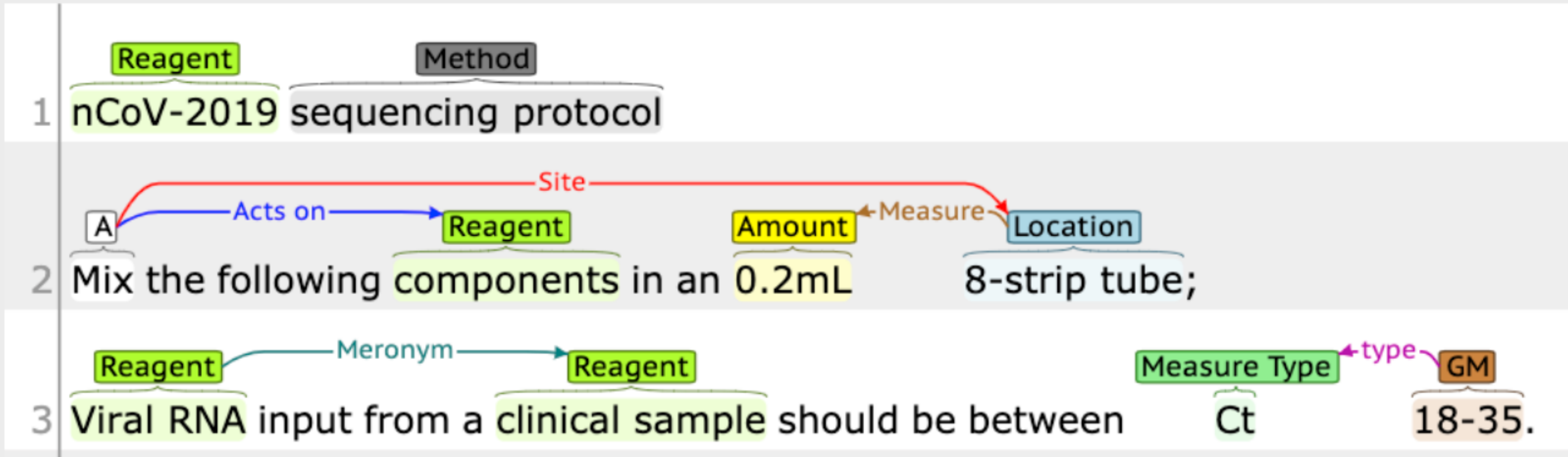}
    \caption{Examples of named entities and relations in a wet lab protocol}
    \label{fig:annotation_interface}
\end{figure}

Wet Lab protocols consist of natural language instructions for carrying out chemistry or biology experiments (for an example, see Figure \ref{fig:annotation_interface}).
While there have been efforts to develop domain-specific formal languages in order to support robotic automation\footnote{\url{https://autoprotocol.org/}} of experimental procedures \cite{bates2017wet}, the vast majority of knowledge about how to carry out biological experiments or chemical synthesis procedures is only documented in natural language texts, including in scientific papers, electronic lab notebooks, and so on.

Recent research has begun to apply human language technologies to extract structured representations of procedures from natural language protocols \cite{kuniyoshi2020annotating, vaucher2020automated, kulkarni2018annotated, soldatova2014exact2, vasilev2011software,  ananthanarayanan2010biocoder}. Extraction of  named entities and relations from these protocols is an important first step towards machine reading systems that can interpret the meaning of these noisy human generated instructions.  

However, performance of state-of-the-art tools for extracting named entity and relations from wet lab protocols still lags behind well edited text genres \cite{jiang2020generalizing}. This motivates the need for continued research, in addition to new datasets and tools adapted to this noisy text genre.

In this overview paper, we describe the development and findings of a shared task on named entity and relation extraction from the noisy wet lab protocols, which was held at the 6-th Workshop on Noisy User-generated Text (WNUT 2020) and attracted 15 participating teams.

In the following sections, we describe details of the task including training and development datasets in addition to the newly annotated test data. We briefly summarize the systems developed by selected teams, and conclude with results.

\begin{table*}[hbtp]
\hspace*{-1cm}%
    \centering
    \begin{subtable}{.6\linewidth}
      \centering
        
        \footnotesize
            \begin{tabular}{l|rrrr|r}
            \hline
             & \textbf{ Train} & \textbf{Dev}   & \textbf{Test-18}  & \textbf{Test-20} & \textbf{Total} \\
            \hline
            \#protocols        & 370   & 122   & 123 & 111   & 726   \\
            
            \#sentences        & 8444    &  2839 & 2813  & 3562  &  17658 \\
            \#tokens           & 107038 & 36106 & 36597 & 51688 & 231429 \\
            \#entities         &  48197 & 15972 & 16490 & 104654 & 185313 \\
            \#relations        & 32158 & 10812 & 11242 & 70591 & 124803 \\
            \hline
            \end{tabular}
            
            \label{tab:corpus_data_stat_total}
    \end{subtable}%
    \begin{subtable}{.4\linewidth}
      \centering
       \footnotesize
            \begin{tabular}{p{.8 in}|r|r}
            \hline
              & \textbf{ per  Protocol} & \textbf{ per  Sentence} \Bstrut\\ 
            \hline
            avg. \#sentences     &   24.32  & - \Tstrut\\[.5 mm]
            avg. \#tokens         & 318.77 & 13.11 \\[.5 mm]
            avg. \#entities     & 255.25  & 10.49 \\[.5 mm]
            avg. \#relation & 171.90  &  7.07 \\[.8 mm]
            \hline
            \end{tabular}
            \label{tab:corpus_data_stat_avg}
    \end{subtable} 
    \caption{Statistics of the Wet Lab Protocol corpus.}
\label{tab:corpus_data_stat}
\end{table*}

\section{Wet Lab Protocols}

Wet lab protocols consist of the guidelines from different lab procedures which involve chemicals, drugs, or other materials in liquid solutions or volatile phases.  The protocols contain a sequence of steps that are followed to perform a desired task. These protocols also include general guidelines or warnings about the materials being used. The publicly available archive of \url{protocol.io} contains such guidelines of wet lab experiments, written by researchers and lab technicians around the world.  This protocol archive covers a large spectrum of experimental procedures including neurology, epigenetics, metabolomics, stem cell biology, etc.  Figure \ref{fig:annotation_interface} shows a representative wet lab protocol.

The wet lab protocols, written by users from all over the worlds, contain domain specific jargon as well as numerous nonstandard spellings, abbreviations, unreliable capitalization. Such diverse and noisy style of user created protocols imposed crucial challenges for the entity and relation extraction systems. Hence, off-the-shelf named entity recognition and relation extraction tools, tuned for well edited texts, suffer a severe performance degradation when applied to noisy protocol texts \cite{kulkarni2018annotated}. 

To address these challenges, there has been an increasing body of work on adapting entity and relation extraction  recognition tools for noisy wet lab texts \cite{jiang2020generalizing, luan-etal-2019-general, kulkarni2018annotated}. However, different research groups have used different evaluation setups (e.g., training / test splits) making it challenging to perform direct comparisons across systems. By organizing a shared evaluation, we hope to help establish a common evaluation methodology (for at least one dataset) and also promote research and development of NLP tools for user generated wet-lab text genres.

\subsection{Annotated Corpus }

Our annotated wet lab corpus includes 726 experimental protocols from the 8-year archive of ProtocolIO (April 2012 to March 2020). These protocols are  manually annotated with 15 types of relations among the 18 entity types\footnote{Our  annotated corpus is available at: {\url{https://github.com/jeniyat/WNUT_2020_NER}}.}. The fine-grained entities can be broadly classified into 5 categories:  {\sc Action},  {\sc Constituents},  {\sc Quantifiers}, {\sc Specifiers}, and  {\sc Modifiers}.  The  {\sc Constituents}  category includes mentions of {\sc Reagent}, {\sc Location}, {\sc Device}, {\sc Mention}, and  {\sc Seal}. The  {\sc Quantifiers}  category includes mentions of {\sc Amount}, {\sc Concentration}, {\sc Size}, {\sc Time}, {\sc Temperature}, {\sc pH}, {\sc Speed}, {\sc Generic-Measure} and  {\sc Numerical}. The  {\sc Specifiers}  category includes mentions of  {\sc Modifier}, {\sc Measure-Type} and  {\sc Method}. The {\sc Action} entity  refers to the phrases denoting tasks that are performed to complete a step in the protocol. The mentions of these entities contain different types of relations, including-- {\sc Site}, {\sc Setting}, {\sc Creates}, {\sc Measure-Type-Link}, {\sc Co-reference-Link}, {\sc Mod-Link}, {\sc Count}, {\sc Meronym}, {\sc Using}, {\sc Measure}, {\sc Commands}, {\sc Of-Type}, {\sc Or},  {\sc Product}, and  {\sc Acts-on}.

\subsubsection{Train and Development data }

The training and development dataset for our task was taken from previous work on wet lab corpus \cite{kulkarni2018annotated} that consists of from the 623 protocols. We excluded the eight duplicate protocols from this dataset and then re-annotated the 615 unique protocols in BRAT \cite{brat_tool}. This re-annotation process aided us to add the previously missing 20,613 missing entities along with 10,824 previously missing relations and also to facilitate removing the inconsistent annotations. The updated corpus statics is provided in Table \ref{tab:corpus_data_stat}. This full dataset (Train, Dev, Test-18) was provided to the participants at the beginning of the task and they were allowed to use any of part of this dataset to train their final model.

\subsubsection{Test Data}

For this shared task we added 111 new protocols (Test-20) which were used to evaluate the submitted models. Test-20 dataset consists of 100 randomly sampled general protocols and 11 manually selected covid-related protocols from ProtocolIO (\url{https://www.protocols.io/}).  This 111 protocols were double annotated by three annotators using a web-based annotation tool, BRAT \cite{brat_tool}. Figure \ref{fig:annotation_interface} presents a screenshot of our annotation interface. We also provided the annotators a set of guidelines containing the entity and relation type definitions.  The annotation task was split in multiple iterations. In each iteration, an annotator was given a set of 10 protocols. An adjudicator then went through all the entity and relation annotations in these protocols and resolved the disagreements. Before adjudication, the inter-annotator agreement is 0.75 , measured by Cohen’s Kappa \cite{cohen_kappa}.

\subsection{Baseline Model}

We provided the participants baseline model for both of the subtasks. The baseline model for named entity recognition task utilized a feature-based CRF tagger developed using the CRF-Suite\footnote{\url{http://www.chokkan.org/software/ crfsuite/}} with a standard set of contextual, lexical and gazetteer features. The baseline relation extraction system employed a feature-based logistic regression model developed using the Scikit-Learn\footnote{\url{https://scikit-learn.org/}} with a standard set of contextual, lexical and gazetteer features.

\begin{center}

\begin{table*}[hbt!]

\hspace*{-1 cm}%
\centering
\footnotesize
\addtolength{\tabcolsep}{-0.05 in} 
\begin{tabular}{p{0.88 in}|p{1.8 in}p{1.6 in}p{1.4 in}}

\hline
Team & Word Representation & Features & Approach\\
    \hline
    
    BiTeM & BERT, BioBERT, RoBERTa, XLNet & - & Ensemble of Transformers\\
    PublishInCovid19 & PubMedBERT & - & Ensemble of BiLSTM-CRFs \\
    
    Fancy Man  & BERT & - & BERT fine tuning \\
    mahab & BERT & Lexical  & BERT fine tuning \\
    mgsohrab & SciBERT & Lexical  & SciBERT fine tuning\\
    SudeshnaTCS & XLNet & Rules & XLNet fine tuning\\
    IITKGP  & BioBERT & - & BioBERT fine tuning \\

    B-NLP & SciBERT, word2vec & - & Biaffine Classifier  \\
    BIO-BIO & BioBERT & - & BiLSTM-CRF\\
    DSC-IITISM  & GLoVe, CamemBERT, Flair & - & BiLSTM-CRF \\
    Kabir  & GLoVe, ELMo, BERT, Flair & Gazetteers & RNN-CRF\\
    
    IBS  & - & Gazetteers, POS Tagger & Ensemble of CRFs \\
    KaushikAcharya  & - & POS Tagger, Dependency Parser & CRF \\
    Baseline  & - & Gazetteers, Lexical, Contextual  & CRF \\

    \hline
\end{tabular}

\caption{Summary of NER systems designed by each team.}
\label{tab:team_model}
\end{table*}
\end{center}

\begin{table}[h!]
\centering
\footnotesize
\begin{tabular}{p{0.85 in}|p{1.8 in}}
\hline
Team Name & Affiliation \\
    \hline
    B-NLP & Bosch Center for Artificial\\
          & Intelligence \\[0.3 ex]
    Big Green & Dartmouth College \\[0.3 ex]
    BIO-BIO & Harbin Institute of technology, \\
            & Shenzhen \\[0.3 ex]
    \multirow{3}{*}{BiTeM} & University of Applied Sciences and Arts of Western Switzerland, Swiss Institute of Bioinformatics, \\
    & University of Geneva \\[0.3 ex]
    
    DSC-IITISM & IIT(ISM) Dhanbad \\[0.3 ex]
    \multirow{3}{*}{Fancy Man} & University of Manchester, Xian Jiaotong University, East China University of Science and Technology,\\ 
    & Zhejiang University  \\[0.3 ex]
    
    IBS & IBS Software Pvt. Ltd, NTNU \\[0.3 ex]
    IITKGP  & IIT, Kharagpur \\[0.3 ex]
    
    Kabir & Microsoft \\[0.3 ex]
    KaushikAcharya & Philips \\[0.3 ex]
    mahab & Amirkabir University of Technology \\[0.3 ex]
    \multirow{2}{*}{mgsohrab} & National Institute of Advanced   \\
    & Industrial Science and Technology \\[0.3 ex]
    PublishInCovid19 & Flipkart Private Limited \\[0.3 ex]
    SudeshnaTCS & TCS Research \& Innovation Lab \\[0.3 ex]
    
    \hline
\end{tabular}
\caption{Team Name and affiliation of the participant.}
\label{tab:team_affl}

\end{table}

\subsection{NER Systems}

Thirteen teams (Table \ref{tab:team_affl}) participated in the named entity recognition sub-task. A wide variety of approaches were taken to tackle this task. Table \ref{tab:team_model} summarizes the word representations, features and the machine learning approaches taken by each team.  Majority of the teams (11 out of 13) utilized contextual word representations. Four teams combined the contextual word representations with global word vectors. Only two teams did not use any type of word representations and relied entirely on hand-engineered features and a CRF taggers. The best performing teams utilized a combination of contextual word representation with ensemble of learning. Below we provide a brief description of the approach taken by each team.

\paragraph{{\textbf{B-NLP}} \cite{B-NLP}} modeled the NER as a parsing task and uses a biaffine classifier. The second classifier of their system used the predictions from the first classifier and then updated the labels of the predicted entities. Both of the classifiers utilized word2vec \cite{mikolov2013efficient} and SciBERT \cite{Lee_2019} word representations.

\paragraph{{\textbf{BIO-BIO}} \cite{BIO-BIO}} implemented a BiLSTM-CRF tagger that utilized BioBERT \cite{lee2020biobert} word representation.

\paragraph{{\textbf{BiTeM}} \cite{BITEM}} developed a voting based ensemble classifier containing 14 transformer models, and utilized 7 different word representations including BERT \cite{devlin2018bert}, ClinicalBERT \cite{huang2019clinicalbert}, PubMedBERT\textsubscript{base} \cite{gu2020domain}, BioBERT \cite{lee2020biobert}, RoBERTa \cite{liu1907roberta}, Biomed-RoBERTa\textsubscript{base} \cite{gururangan2020don} and XLNet \cite{yang2019xlnet}.

\paragraph{{\textbf{DSC-IITISM}} \cite{DSC-IITISM}} developed a BiLSTM-CRF model that utilized a  concatenation of CamemBERT\textsubscript{base} \cite{martin-etal-2020-camembert}, Flair(PubMed) \cite{akbik2018coling}, and GloVe(en) \cite{pennington2014glove} word representations.

\paragraph{{\textbf{Fancy Man}} \cite{Fancy-Man}} fine-tuned the  BERT\textsubscript{base} \cite{devlin2018bert} model with an additional linear layer.

\paragraph{{\textbf{IBS}} \cite{IBS}} utilized an ensemble classifier with 4 feature based on CRF taggers.

\paragraph{{\textbf{Kabir}} \cite{kabir}} employed an RNN-CRF model that  utilized concatenation of Flair(PubMed) \cite{akbik2018coling} and ELMo(PubMed) \cite{Peters:2018} word representations.

\paragraph{{\textbf{KaushikAcharya}} \cite{KaushikAcharya}} employed a linear CRF with hand-crafted features.

\paragraph{{\textbf{mahab}} \cite{mahab}} fine-tuned the BERT\textsubscript{base} \cite{devlin2018bert} sequence tagging model.

\paragraph{{\textbf{mgsohrab}} \cite{mgsohrab}} fine-tuned  the SciBERT \cite{beltagy2019scibert} model.

\paragraph{{\textbf{PublishInCovid19}} \cite{PublishInCovid19}}  employed a structured ensemble classifier \cite{nguyen2007comparisons} consisting of 11 BiLSTM-CRF taggers, that utilized the PubMedBERT \cite{gu2020domain} word representation.

\paragraph{{\textbf{SudeshnaTCS}} \cite{SudeshnaTCS}} fine-tuned  XLNet \cite{yang2019xlnet} model.

\paragraph{{\textbf{IITKGP}} \cite{iitkgp}}  fine-tuned the Bio-BERT \cite{lee2020biobert} model.

\begin{table}[tb!]
\footnotesize
\centering
\addtolength{\tabcolsep}{-0.05 in} 
\begin{tabular}{l|p{.28 in} p{.28 in} p{.28 in}}

\hline
&  \textbf{ \hspace{.02 in} P} &  \textbf{\hspace{.02 in} R}    & \textbf{\hspace{.02 in} F$_{1}$} \\ 
\hline 
\multicolumn{4}{l}{\textit{\textbf{Exact Match}}}   \\ \hline 

BiTeM & \textbf{84.73}  &  72.25 &  \textbf{77.99} \\
PublishInCovid19 &  81.36 & \textbf{74.12} &  \textul{77.57} \\

Fancy Man & 76.21 & 71.76 & 73.92 \\
mahab &   50.19 & 52.96 & 51.54 \\
mgsohrab &  \textul{83.69} & 70.62 &  76.60 \\
SudeshnaTCS &  74.99 & 71.43 &  73.16 \\
IITKGP &  77.00 & \textul{72.93} & 74.91 \\

B-NLP &  77.95 &  63.93 & 70.25 \\
BIO-BIO &  78.49 &  71.06 & 74.59 \\
DSC-IITISM & 64.20 & 57.07 &  60.42 \\
Kabir & 78.79 & 72.20  &75.35 \\

IBS &   74.26 & 62.55 & 67.90 \\
KaushikAcharya & 73.68 & 63.98  & 68.48 \\[0.2 ex]
Baseline & 70.06 & 61.91 &  65.73 \\
\hline
Ensemble & 78.35 & 75.39 &  76.84 \\
\hline

\multicolumn{4}{l}{\textit{\textbf{Partial Match}}}   \\ \hline

BiTeM & \textbf{88.72}  & 75.66 & \textul{81.67} \\
PublishInCovid19  & 85.74 & \textbf{78.11}  & \textbf{81.75} \\

Fancy Man   & 81.15 & 76.41 & 78.71 \\
mahab & 55.09 & 58.14 & 56.57 \\
mgsohrab  & \textul{87.95}  & 74.22 & 80.50 \\
SudeshnaTCS & 79.73 & 75.95 & 77.80 \\
IITKGP  & 81.76 & \textul{77.43}  & 79.54 \\

B-NLP & 84.85 & 69.59 & 76.46 \\
BIO-BIO & 83.16 & 75.29 & 79.03 \\
DSC-IITISM  &   68.52 & 60.90 & 64.49 \\
Kabir & 83.73 & 76.73 & 80.08 \\

IBS & 79.72 & 67.15 & 72.89 \\
KaushikAcharya  & 79.31 & 68.87 & 73.73 \\[0.2 ex]
Baseline & 75.66 & 66.85 &  70.98 \\
\hline
Ensemble & 84.16  & 79.15 & 81.58 \\
\hline

\hline
\end{tabular}

\caption{Results on extraction of 18 Named Entity types from the \textit{Test-20} dataset. \textbf{Exact Match} reports the performance  when the  predicted  entity type is same as the gold entity and the predicted entity boundary is the exact same as the gold entity boundary. \textbf{Partial Match} reports the performance  when the  predicted  entity type is same as the gold entity and the predicted entity boundary has some overlap with gold entity boundary. }
\label{tab:main_results_ner}

\end{table}

\begin{figure}[tb!]
    \centering
    \hspace*{-2 mm}
    \includegraphics[scale=0.21]{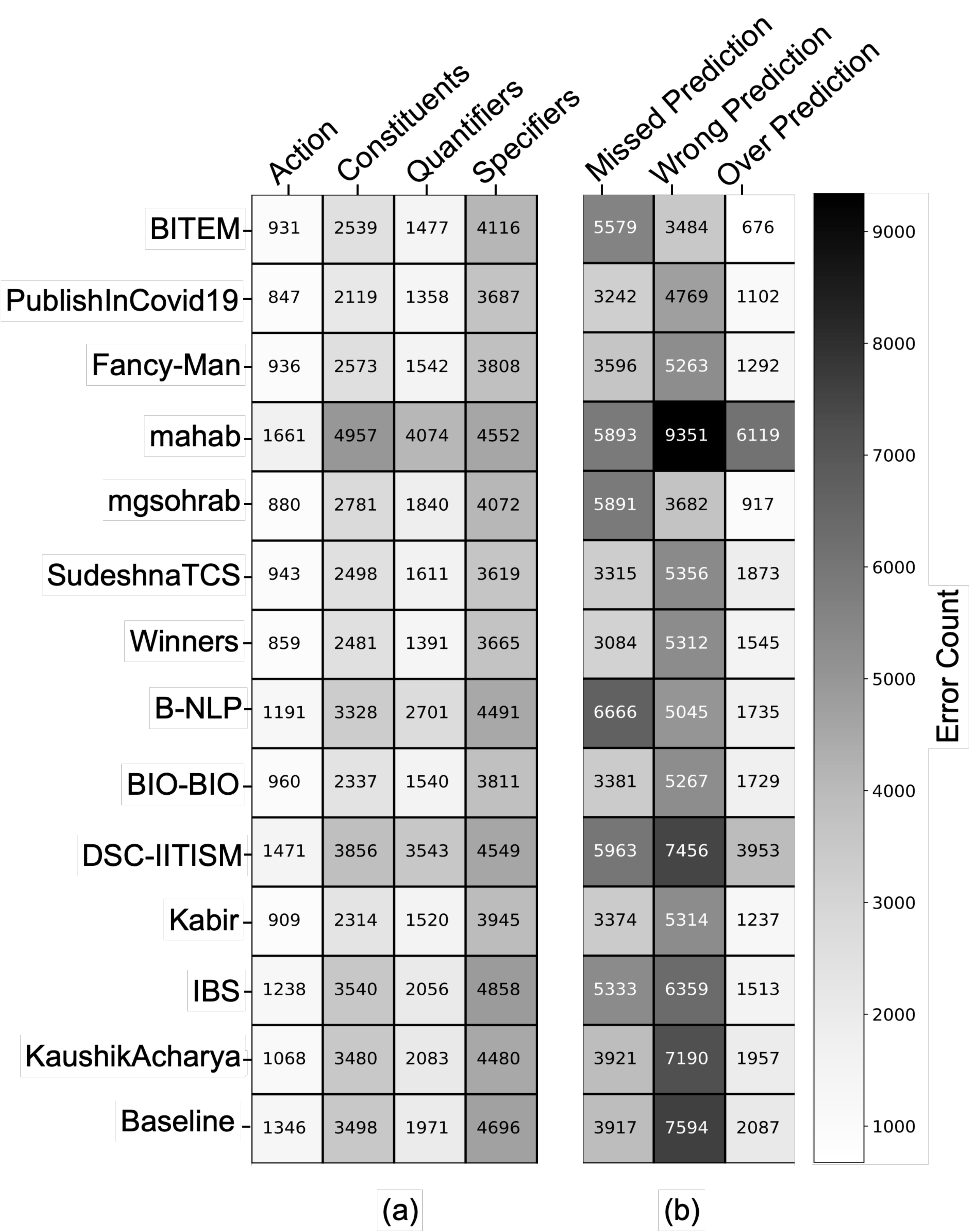}
    \caption{Summary of incorrectly classified entity tokens by each submitted systems.}
    \label{fig:error_analysis}
\end{figure}

\subsection{RE Systems}

Two teams (Table \ref{tab:team_affl}) participated in the relation extraction sub-task. Both of the teams followed fine-tuning of contextual word representation and did not use any hand-crafted features. Table \ref{tab:team_model_re} summarizes the word representations and the machine learning approaches followed by each team. Below we provide a brief description of the model developed by taken by each team.

\paragraph{{\textbf{Big Green}} \cite{BigGreen}}  considered the protocols as a knowledge graph, in which relationships between entities are edges in the knowledge graph. They trained a BERT \cite{devlin2018bert} based system to classify edge presence and type between two entities, given entity text, label, and local context.

\paragraph{{\textbf{mgsohrab} \cite{mgsohrab}}} utilized PubMedBERT \cite{gu2020domain} as input to the relation extraction model that enumerates all possible pairs of arguments using deep exhaustive span representation approach.

\begin{table*}[hbt!]
\hspace*{-1 cm}%
\centering
\footnotesize
\addtolength{\tabcolsep}{-0.05 in} 
\begin{tabular}{p{0.88 in}|p{1.8 in}p{1.6 in}p{1.4 in}}

\hline
Team & Word Representation & Features & Approach\\
    \hline
    mgsohrab & PubMedBERT & - & PubMedBERT fine-tuning \\
    Big Green & BERT & - & BERT fine-tuning  \\
    Baseline  & - & Gazetteers, Lexical, Contextual  & Logistic Regression \\
    
    \hline
\end{tabular}

\caption{Summary of relation extraction systems designed by each team.}
\label{tab:team_model_re}
\end{table*}

\section{Evaluation}
In this section, we present the performance of each participating systems along with a description of the errors made by the model types.

\subsection{NER Errors Analysis}

Table \ref{tab:main_results_ner} shows the comparison of precision (\textbf{P}), recall (\textbf{R}) and \textbf{F$_1$} score among different teams, evaluated on the {\em{Test-20}} corpus. Here, the exact match refers to the case where  predicted entity type and  boundary are  exactly same as the gold entity type and  boundary. Whereas, the partial match refers to the case where the predicted entity type is  same as the gold entity type and predicted entity boundary has some overlap with the gold entity boundary.

We observe that ensemble models with contextual word representations outperforms all other approaches by achieving 77.99 F\textsubscript{1} score in exact match (Team:BITEM) and 81.75 F\textsubscript{1} score in partial match (Team:PublishInCovid19). 

In Figure \ref{fig:error_analysis}, we present an error analysis of different NER systems.

Analysis of the errors these different NER model prediction demonstrate that, the BERT based models make less mistakes in false positive and incorrect type errors compared to the traditional neural networks and feature based models. We also observed that, these BERT models suffer from higher false negatives errors compared to the other approaches. 

To combine the advantages of these different approaches, we made an majority voting based ensemble classifier. Our ensemble NER tagger utilizes the predictions of all the submitted systems and then it assigns each word the most frequently predicted tag. This ensemble classifier performs better than all the single fine-tuned BERT models and it outperformed the traditional neural and feature based models by achieving 76.84 F\textsubscript{1} (Table \ref{tab:main_results_ner}). However, our ensemble NER tagger performed 1.15 F\textsubscript{1} below the neural ensemble models (Team:BITEM, PublishInCovid19). We would like to note that,  we did not have access to the participant model’s predictions on development and training set. Hence, it was not possible for us to fine-tune our ensemble classifier on the entity recognition task.

\subsection{RE Errors Analysis}
\begin{figure}[h!]
    \centering
    \hspace*{-2 mm}
    \includegraphics[scale=0.3]{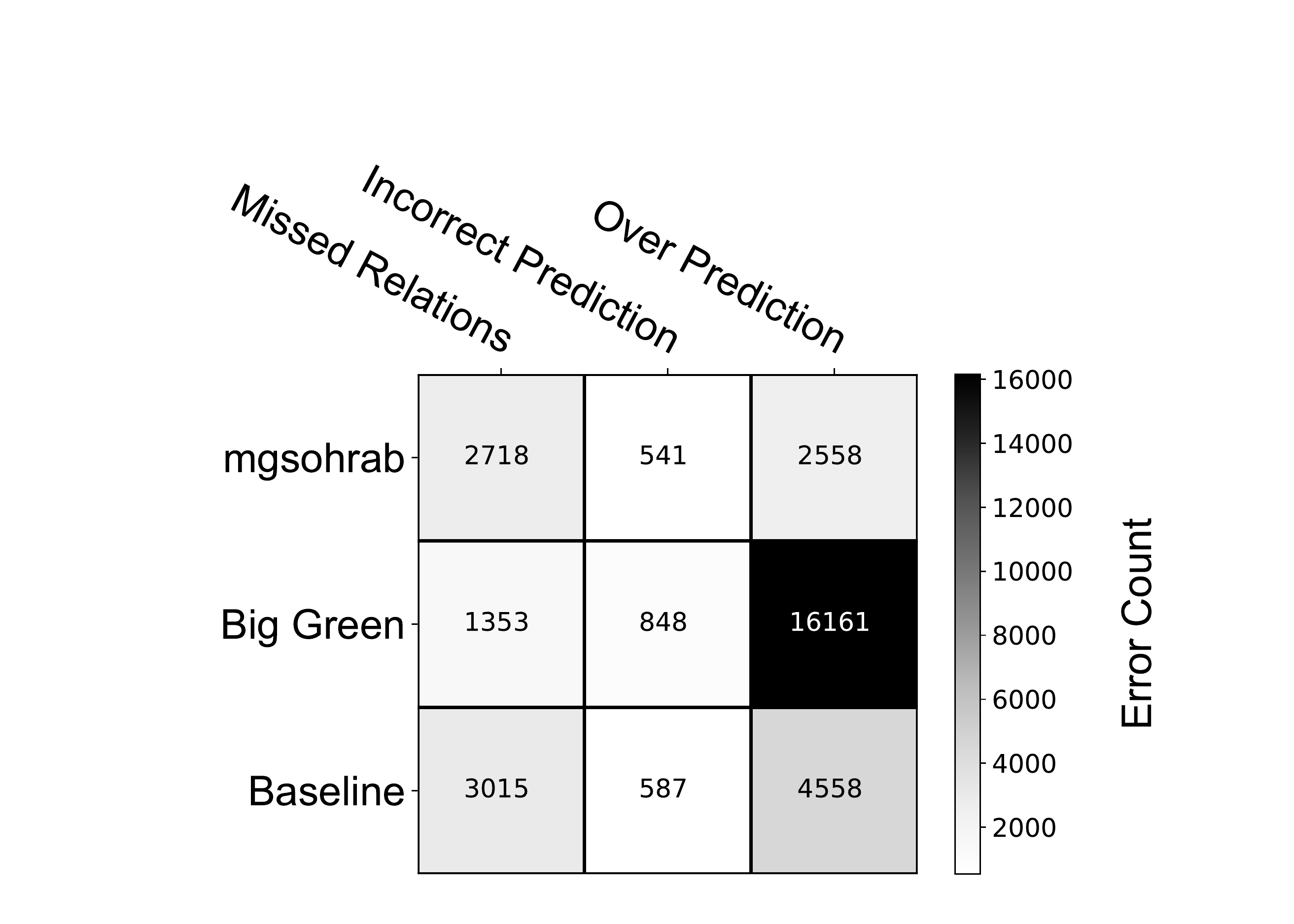}
    \caption{Summary of incorrectly predicted relations in each submitted systems. }
    \label{fig:error_analysis_rel}
\end{figure}

Table \ref{tab:main_results_re} shows the comparison of precision (\textbf{P}), recall (\textbf{R}) and \textbf{F$_1$} score among the participating teams, evaluated on the {\em{Test-20}} corpus. Both of the teams utilized the gold entities and then predicted the relations among these entities by fine-tuning contextual word representations. We observed that fine-tuning of domain related PubMedBERT provides significantly higher performance compared to the general BERT fine-tuning. While examining the relation predictions from both of these systems, we found that system with PubMedBERT fine-tuning (Team:mgsohrab) resulted in significantly less amount of errors in every category (Figure \ref{fig:error_analysis_rel}).

\begin{table}[h!]
\footnotesize
\centering
\addtolength{\tabcolsep}{-0.05 in} 
\begin{tabular}{l|p{.28 in} p{.28 in} p{.30 in}}

\hline
&  \textbf{ \hspace{.02 in} P} &  \textbf{\hspace{.02 in} R}    & \textbf{\hspace{.02 in} F$_{1}$} \\ 
\hline

mgsohrab & \textbf{80.86} & \textul{80.07}   & \textbf{80.46}\\
Big Green & 45.42 & \textbf{86.54} &  59.57\\
Baseline & \textul{80.10} & 66.21 & \textul{72.50} \\
\hline
Ensemble & 82.65  & 80.32 & 81.32 \\
\hline

\end{tabular}

\caption{Results on extraction of 15 relation types from the \textit{Test-20} dataset.}
\label{tab:main_results_re}

\end{table}

The error analysis over different participant predictions revealed that the general domain BERT has less false negative errors compared to the domain-related BERT. However, the domain related Pubmed-BERT models have significantly less number of  false positive and incorrect type errors compared to the general domain BERT. To combine the advantages of these different approaches, we made an ensemble classifier from the prediction of the submitted systems, where we assign the most frequently predicted relation for each entity pair. This ensemble classifier outperforms the winner system by achieving 81.32 F\textsubscript{1} score.

\section{Related Work}
The task of information extraction from wet lab protocols is closely related to the event trigger extraction task. The event trigger task has been studied extensively, mostly using ACE data \cite{doddington2004automatic} and the BioNLP data \cite{nedellec2013overview}. Broadly, there are two ways to classify various event trigger detection models: (1) \textit{Rule-based} methods using pattern matching and regular expression to identify triggers \cite{vlachos2009biomedical} and (2) \textit{Machine Learning based} methods focusing on generation of high-end hand-crafted features to be used in classification models like SVMs or maxent classifiers \cite{pyysalo2012event}. Kernel based learning methods have also been utilized with embedded features from the syntactic and semantic contexts to identify and extract the biomedical event entities \cite{zhou2014event}. In order to counteract highly sparse representations, different neural models were proposed. These neural models utilized the dependency based word embeddings with feed forward neural networks \cite{wang2016emb}, CNNs \cite{wang2016cnn} and Bidirectional RNNs \cite{rahul2017biomedical}. 

Previous work has experimented on datasets of well-edited biomedical publications with a small number of entity types. For example, the JNLPBA corpus \cite{kim2004introduction} with 5 entity types (\verb!CELL LINE!, \verb!CELL TYPE!, \verb!DNA!, \verb!RNA!, and \verb!PROTEIN!) and the BC2GM corpus \cite{hirschman2005overview} with a single entity class for genes/proteins. In contrast, our dataset addresses the challenges of recognizing 18 fine-grained named entities along with 15 types of relations from the user-created wet lab protocols.

\section{Summary}
In this paper, we presented a shared task for consisting of two sub-tasks: named entity recognition and relation extraction from the wet lab protocols.  We described the task setup and datasets details, and also outlined the approach taken by the participating systems. The shared task included larger and improvised dataset compared to the prior literature \cite{kulkarni2018annotated}. This improvised dataset enables us  to draw stronger conclusions about the true potential of different approaches. It also facilitates us in analyzing the results of the participating systems, which aids us in suggesting potential research directions for both future shared tasks and noisy text processing in user generated lab protocols.

\section*{Acknowledgement}

We would like to thank Ethan Lee and Jaewook Lee for helping with data annotation.  This material is based upon work supported by the Defense Advanced Research Projects Agency (DARPA) under Contract No. HR001119C0108.  The views, opinions, and/or findings expressed are those of the author(s) and should not be interpreted as representing the official views or policies of the Department of Defense or the U.S. Government.

\bibliographystyle{acl_natbib}
\bibliography{references}

\end{document}